\documentclass[preprint,12pt,3p]{elsarticle}

\usepackage{amssymb}
\usepackage{amsmath}
\usepackage{gensymb}
\usepackage{rotating}
\usepackage{tikz}
\usepackage{algorithm}
\usepackage{caption}
\usepackage{subcaption}
\usepackage{multirow}
\usepackage[noend]{algpseudocode}
\usepackage{vruler}
\usepackage{fancyhdr,lipsum}

\makeatletter
\def\BState{\State\hskip-\ALG@thistlm}
\makeatother

\begin{document}

\begin{frontmatter}

\title{Optimal Reservoir Operations using Long Short-Term Memory Network}





\author[label1]{Asha Devi Singh\corref{cor1}}
\address[label1]{G.B Pant Institute of Technology, New Delhi, India}

\cortext[cor1]{Corresponding Author: Asha Devi Singh}
\ead{adsingh@gbpit.ac.in}

\address[label5]{Netaji Subhas Institute of Technology, University of Delhi,India}

\author[label5]{Anurag Singh}
\ead{anurags.it@nsit.net.in}

\begin{abstract}
Reliable forecast of inflows to the reservoir is a key factor in the optimal operation of reservoirs. Real-time operation of the reservoir based on forecasts of inflows can lead to substantial economic gains. However, the forecast of inflow is an intricate task as it has to incorporate impacts of climate and hydrological changes. Therefore, the major objective of the present work is to develop a novel approach based on long short-term memory (LSTM) for the forecast of inflows. Real-time inflow forecast, in other words, daily inflow at the reservoir helps in efficient operation of water resources. Also, daily variations in the release can be monitored efficiently and reliability of operation is improved. This work proposes a naive anomaly detection algorithm baseline based on LSTM. In other words, a strong baseline to forecast flood and drought for any deep learning-based prediction model. The practicality of the approach has been demonstrated using the observed daily data of the past 20 years from Bhakra Dam in India. The results of the simulations conducted herein clearly indicate the supremacy of the LSTM approach over the traditional methods of forecasting. Although, experiments are run on data from Bhakra Dam Reservoir in India, LSTM model and anomaly detection algorithm are general purpose and can be applied to any basin with minimal changes. A distinct practical advantage of the LSTM method presented herein is that it can adequately simulate non-stationarity and non-linearity in the historical data.
\end{abstract}
\begin{keyword}
LSTM, Thomas-Fiering model, Inflow forecasting, RMSE
\end{keyword}

\end{frontmatter}


\section{Introduction}
\label{sec1}
All forms of life on earth depend on water. The distribution of water is, however, not uniform. It varies both spatially and temporally across regions. At certain locations, ample water is available, whereas some areas face scarcity of water and are frequently subjected to drought. Areas having abundant water face challenges of managing it and preventing the area from flood. For efficient use of river water, dams are built and management of water resource is carried out. Due to climate change exhaustive planning of water resource is required to meet the water demands in the dry season and firm power demand under stress of increased climate variability.
\par Optimal operation of reservoir system is essential to meet competing water demands. Various techniques to achieve optimal operation of reservoir system have been used to treat non-convex, non-linear behavior of reservoir systems. Dynamic Programming (DP) \cite{bellman1966dynamic}(Bellman, 1966) is the widely accepted technique for optimization of reservoir system as these are characterized by a large number of non-linear and stochastic features that can be translated into DP formulation. Thomas-Fiering model is a widely used technique to forecast inflows, and can be effectively used for deciding reservoir release policies.  \cite{sargent1979simplified}(Sargent,1979) used transition probabilities to generate sequences of daily stream flows while preserving the important characteristics of the daily inflow hydrograph. In the past a wide variety of approaches have been used to forecast inflows. In this paper a novel long short-term memory (LSTM) based neural network architecture has been developed for inflow forecasting.
\par 
The use of LSTM to forecast inflows has not been attempted to the best of our knowledge. Considerable success has been achieved with the use of machine learning techniques which are able to solve wide range of problems in optimization and operations research \cite{singh2019image,singh2020image}(Singh, 2019, 2020)  \cite{dutta2020adaptive,dutta2020styleguide}(dutta 2020). Recently attempts have been made for successful determination of precipitation but naive deep learning based architectures are difficult to train them to learn and investigate temporal correlation over arbitrary length. Artificial Neural Networks (ANN) are essentially functions that consist of large set of parameters i.e. weights that try to fit the data by tuning them. ANNs does not have  any  consideration of  the temporal sequence  within the data. Several applications of ANNs to water resource management problems have been reported in the literature \cite{zealand1999short} (Zealand et.al,1999). On the other hand, recurrent neural networks (RNNS) have self referencing feedback loop in their architecture. The use of recurrent neural networks has been proposed by several researchers \cite{coulibaly2001multivariate} (Coulibaly et.al, 2017). When considered theoretically, RNNs are capable of learning to track relationships for arbitrary lengths in temporal input. However, it becomes intractable to keep account of learning for arbitrary length. RNN fails as gradients being used for computation cannot keep values for arbitrary precision which then turn to zero or explode.
\par 
Precipitation and inflow prediction are essentially time series problems and have a temporal correlation in the data. Since the approximate period for such repetition will be no less than an  year based upon first principles. We can safely conclude that remembering gradients for such long iterations to take into account data for last year can cause problems in case of RNNs due to their limitations of vanishing gradients. RNN could be employed for cases when month wise average inflow needs to be computed for the reservoir system. Monthly average inflow predictions cannot reveal much significant information for design of strategies for daily operation of reservoir, therefore not serving much fruition in real use case. The Long Short Term network is a recurrent neural network which is trained using back propagation through time and also overcomes problem of vanishing gradient. 
\par
Anomaly in daily inflow prediction by LSTM will also help in prediction of drought and flood. Result obtained using LSTM network reveals that model is satisfactorily able to stimulate non stationary and non- linear inflow trends.

\section{Literature Review}
\label{sec2}
Hydrological process are intricate as it is difficult to understand the complex underlying process that generate the observed system dynamics. \cite{thomas1962mathematical}(Thomas, H.A et.al 1962) made use of Auto regressive model to generate inflow assuming hydrological data as time series and stochastic in nature. \cite{singhal1980mathematical}(M.Singhal, et.al,1980)  developed a mathematical model using Thomas-Fiering model suggest that inflow in any month is dependent on previous month and also depends on the inflow of the previous year of same month. Model was used for Matatila dam on river Betwa  \cite{faruk2010hybrid}(D.O Faruk,2010) has developed a model combining ARIMA and neural network. ARIMA model is unable to deal with non-linear relationship where as neural network model alone is incapable of handling  linear and non-linear pattern for accurate estimation of time series data. The  hybrid model was tested for 108-month  observations  of water quality and the results obtained  were promising. In \cite{khandelwal2015time} (I. Khandelwal et. al, 2015) proposed a novel technique of forecasting by segregating a time series dataset into linear and non-linear components. Thereafter, the Autoregressive Integrated Moving Average (ARIMA) was used to predict linear component and the ANN model was used to predict non linear components. Authors have used the strength of Discrete wavelet Transform, ARIMA, and ANN to improve the accuracy of forecasts.
\par
In water resource planning and management streamflow forecasting plays critical role. \cite{sun2014monthly} (Sun et.al, 2015) described a streamflow forecasting approach using Gaussian Process Regression (GPR), which is an effective kernel-based machine learning algorithm. Results obtained using GPR are more promising when compared to linear regression and artificial neural network models. 
\cite{stedinger1984stochastic}(J.R.Stedinger et.al,1984) developed a stochastic dynamic programming model to forecast the current period inflow for devising reservoir release policy. Expected benefits from future operations using forecast inflows are also estimated. \cite{mujumdar2007bayesian}(Mujumdar et.al,2007) developed an operating policy for the Kalindi Hydroelectric Project Stage-1 of the state of Karnataka in India using a Bayesian stochastic dynamic programming model. The performance of the model is measured by estimating its deviation from the total firm power.
\cite{raso2017effective}(Raso et.al,2007) developed a stochastic dual dynamic programming model for the determination of stream flows. Based on generated streamflow, reservoir operating rules were decided for Menantali Reservoir located on the river Senegal, West Africa. \cite{li2010dynamic}(Li et.al. 2010) used dynamic programming to estimate stochastic inflows considering its uncertainty. The authors have estimated future inflows from the available records with an assumption that the inflow forecasting error has a normal distribution.
\par 
\cite{philbrick1999limitations}(Philbrick et.al, 1999) proposed the application of deterministic optimization for devising reservoir operating policies. The authors concluded that it is possible to solve large scale problems  without much simplification when a reliable forecast of inflow is available.  \cite{naadimuthu1982stochastic}(Naadimuthu et.al, 1982) demonstrated the application of two nonlinear programming techniques – a generalized reduced gradient technique and a gradient projection technique. Each of these techniques was used in conjunction with a Markovian decision process to solve the problem of multipurpose reservoir systems operation.
\cite{fayaed2013reservoir}(Fayaed et.al. 2013) proposed the integration of stochastic dynamic programming and artificial neural network for optimization of reservoir operation.  \cite{ahmad2007optimal}(Rashid et.al. 2007) used stochastic dynamic programming to obtain optimal operating decisions for Dokan reservoir in Iraq.
\par 
Due to the large availability of hydrological data and increase of computational capacity,  the statistical models are generally developed to estimate the behavior of observed data. Lot of seminal work involving solutions to the problems in different areas is now being powered by deep learning. For prediction of nonlinear hydrologic processes, ANNs are widely used now days.  Using ANNs, forecasting of stream  flow for short term time horizons is feasible. 
\cite{castelletti2007neuro}[20](Castelletti et.al. 2007) used neuro-dynamic programming for management of multi- purpose reservoir instead of stochastic dynamic programming. \cite{zealand1999short}(Zealand.et.al,1999) compared the performance of ANN with the conventional methods of streamflow forecasting. \cite{coulibaly2001multivariate}(Coulibaly et.al, (2017) investigated the performance of three different types of temporal neural networks for reservoir operations. The best results were provided by the recurrent neural network.
\par
\cite{duonglong}(Duong,et.al. 2019) used long short term memory recurrent neural network technique for determination of monthly rainfall predictions. Based upon the predicted rainfall, the inflows to the reservoir were estimated. \cite{lin2009support}(Le et al, 2019) suggested a use of Long Short-Term Memory (LSTM) neural network model for flood forecasting, used daily discharge and rainfall  as input data. Findings of his study were implemented to forecast  flood on the Da River in Vietnam, where the river flow through many countries and downstream flows (Vietnam) may fluctuate suddenly due to flood discharge from upstream hydroelectric reservoirs. \cite{qi2019decomposition}(Yutao  et al, 2019) used LSTM to forecast inflow and the results obtained illustrate superiority  the average absolute percentage error is reduced to 13.11\%, and the normalized mean square error is reduced by 4\%, the coefficient of determination was increased by 5\%. Model is experimented on the Ankang reservoir in China.
\section{Study Area}
\label{sec3}
\begin{figure}
     \centering
     \includegraphics[width=\textwidth,height=340px]{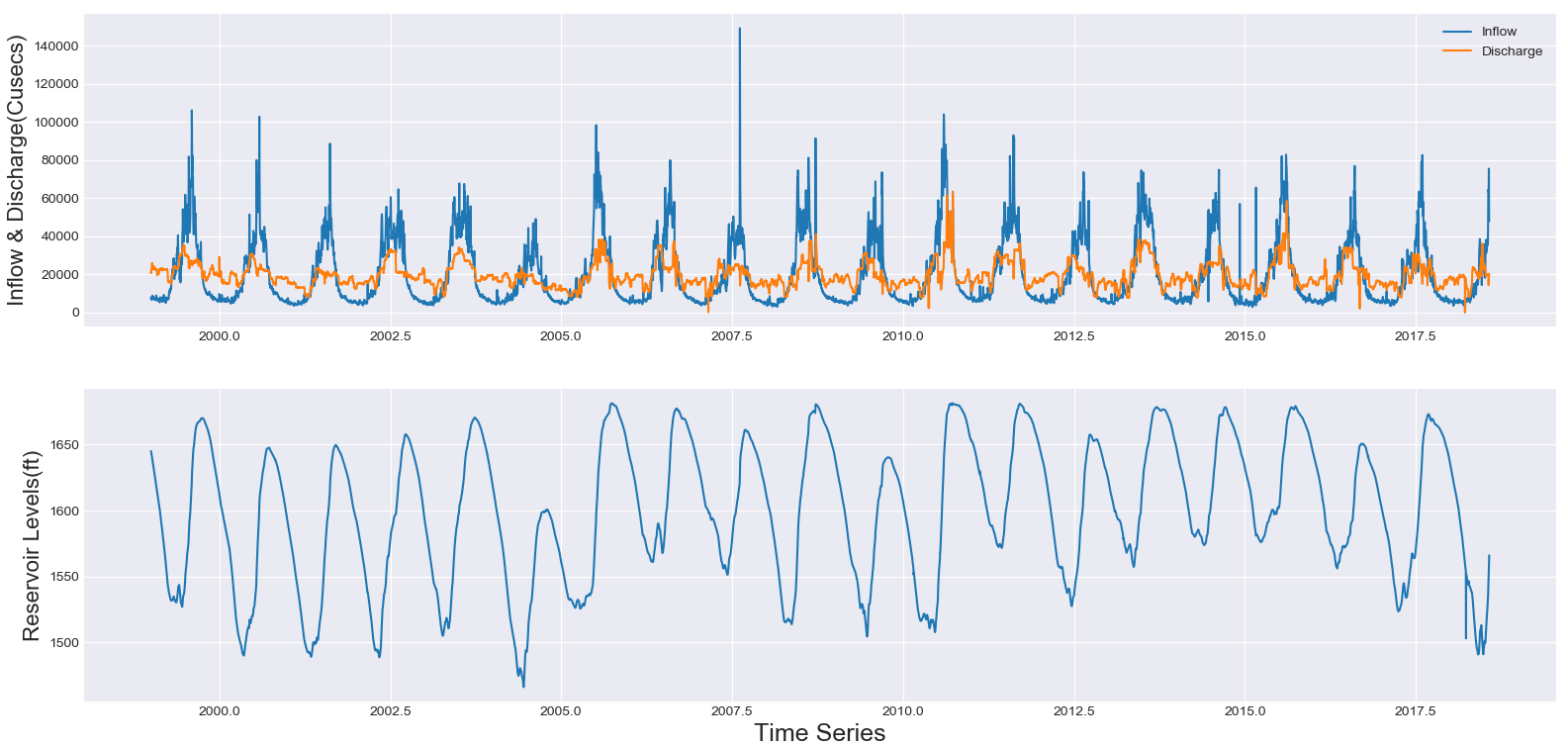}
     \caption{Inflow, Discharge and reservoir levels stats for Bhakra Reservoir.}
     \label{fig:inflow_discharge_levels}
 \end{figure}

Bhakra Dam is located on river Satluj which originates from Mansarowar lake in Tibet at an approximate elevation of 4572 m. The Satluj basin extends from 30\degree N to 33\degree N latitudes and 76\degree E to 83\degree E longitudes. The length of the river is approximately about 1,448 km. River Satluj which originates from Himalayan region provides the critical source of water and is also considered one of the most sensitive area to global  warming.  Due  to  change  in  climate, precipitation pattern changes, causes variation in stream flow. Changes in timing of stream flow even without change in magnitude of stream flows also poses a serious implication for water management. Figure \ref{fig:inflow_discharge_levels} explains in two different sections various observed trends  within the last 20 years of operations in Bhakra Nangal Dam. The first figure explains the relationship between Inflow and discharge in last 20 years using a time series. Second Figure shows trend in reservoir levels on which the dam operates notice the crests and troughs of each  year throughout 20 years, which are spaced evenly indicating strong correlation in months and reservoir levels. Reservoir levels in Figure \ref{fig:inflow_discharge_levels} indicate at the end of filling period reservoir level is El. 1680(Ft).
\par
A major reservoir of Satluj River basin is located at Bhakra. The operation is becoming a complex process due to the large number of uncertainties associated with it, particularly under the influence of climatic changes that have altered precipitation and streamflow patterns in the basin as described by \cite{sharif2013simulation} (Sharif et al. 2013). 
In general, the reservoir operation strategies consist of controlled release of water downstream considering the inflow and the available storage in the reservoir. The objective of operation is to fulfill the demand with respect to power, irrigation, water supply and flood control among other demands.
\par
For the purpose of reservoir operation at Bhakra, an year is divided into two parts. The filling period is from $21^{st}$ May to $20^{th}$ September, whereas the depletion period is from $21^{st}$ September to $20^{th}$ May of the next year. The water accounts are prepared separately for the filling and depletion period. The excess/shortages of one period are not carried over to the next period. The present procedure of meeting the water indents require to supply full water requirements during filling period irrespective of the type of the year (above or below average) expected to encountered.
\par
Depending upon whether the year is going to be wet or a dry one, a suitable reservoir factor is estimated for the depletion period and releases are made accordingly till the reservoir touches down the dead storage level after which only the runoff of the river is passed down. For the above purposes the reservoir factor is defined as a factor by which indents are to be reduced before making the releases and is equal to the ratio:
\begin{equation}
    Reservoir Factor =\frac{Available\hspace{1mm} storage +Total\hspace{1mm} inflow \hspace{1mm}during \hspace{1mm}remaining\hspace{1mm} \hspace{1mm} year
       }{Total \hspace{1mm}water \hspace{1mm}indent\hspace{1mm} during\hspace{1mm}remaining \hspace{1mm}year}
\end{equation}
 where 'Available storage' is water available at the end of filling period, it also considers the reservoir losses during the remaining period of year. Total river flow during the remaining  period of the year is calculated for average 10 daily discharges obtained from the discharge observations for the corresponding 10 day periods for various years for which the discharge record is available, it also includes the effect of losses or gains during the remaining period of the year. The reservoir factor is effective only when its value is 1 or less than 1. Estimating the reservoir factor according to above definition requires estimating the likely inflow during the depletion period. The accuracy of the reservoir factor would thus depend upon the degree of accuracy to which likely inflow during the depletion period can be estimated. 
 Current operational strategies for inflow determination in remaining period depends upon  averages 10 daily of historical data. The rule curve of Bhakra Dam  enumerated below shows that filling of dam is correlated with levels to be attained by certain dates. The RMSE of current strategies is 29.4\% and $R^2$ is 0.6571 . The results clearly indicate variation persists between actual inflow and predicted inflow.

\begin{itemize}
    \item The reservoir should not be filled beyond El. 1650 ft by 31st July
    \item Reservoir level El. 1670 ft should be filled by 15 August not beyond it.
    \item The reservoir level El. 1680 ft should not fill earlier than 31st August.
\end{itemize}
The reservoir should not be filled beyond El. 512.06. (1680ft ) and also this level should not reach earlier than 31st August and  filling above this elevation should  be attempted after ensuring due safeguards.

\section{Dataset}
The observed daily inflow for the period 1999 - 2018 at Bhakra is available on the website of Bhakra Beas Management board (BBMB). The Table \ref{tab:my_label} shows the statistics of the observed daily inflow data at Bhakra for three periods. It is subdivided into Training set, Testing set and validation set as provided in Table \ref{tab:my_label}. It includes minimum, maximum, mean, standard deviation, Kurtosis, Skewness and auto correlation for 1 day lag to 3 day lag( R1, R2, R3) of daily inflow data. Auto correlation indicates high degree of dependency on previous day inflow.
\begin{table}[]
    \centering
    \begin{tabular}{|p{35 mm}|c|c|c|}
    \hline
Parameters& Training Set& Validation Set& Testing Set \\
\hline
Minimum & 3101&	4471&	2767\\
\hline
Maximum &149075&	92890&	131488\\
\hline
Mean&	19226.518&	18079.232&	19754.6\\
\hline
Standard Deviation	&17161.079	&16353.411&	16957.402\\
\hline
Kurtosis&	2.314&	1.875&	1.334\\
\hline
Skewness&	1.486&  1.526&	1.314\\
\hline
R1 &	0.974&	0.969&	0.955\\
\hline
R2 &	0.945&	0.951&	0.936\\
\hline
R3&	0.926&	0.946&	0.918\\
\hline
    \end{tabular}
    \caption{Statistics for Training Set  ,Validation Set and Testing Set of observed daily inflow data}
    \label{tab:my_label}
\end{table}
\section{Flow Forecasting Models}
 \label{5}
Streamflow forecasting techniques are an integral part of real time operation models. Stochastic streamflow models are generally used in simulation studies to assess the response of the water resource systems to future scenarios. The development and implementation of such models require the historical record of flows. Based upon the historical record, the parameters of the rainfall or streamflow forecasting model are generated. The verification of the model is then carried out by comparing the statistics of the generated sequences with the historical sequences. Results indicate relatively strong persistence of streamflow predictability.
\subsection{The Thomas- Fiering Model (TFM)}
  The Thomas-Fiering model (TFM)\cite{thomas1962mathematical}(Thomas, 1962); \cite{Fiering1967streamflow} (Fiering, 1967) is a lag -one Markov model. In most streamflow generation techniques it is sufficient to assume that a first order Markov structure exists. The Thomas Fiering model is fitted to the standardized  monthly flows. 
\begin{equation}
      y_{i,j}=\frac{(x_{i,j}-\bar{x_{j}})}{\sigma_{j}}
\end{equation}
where $x_i,_j$ is the original flow for year $i$ and month $j$ , $\bar{x_{j}}$ and $\sigma_{j}$ are sample estimates of the month and standard deviation respectively, for month $j$.
The model is based on a lag-one auto-regressive Markov process, and the standardized generated flow is given by 
   \begin{equation}
    y_{i,j}=\beta_{j}y_{i-1,j-1}+\epsilon_{i,j}
\end{equation} 
where $\epsilon_{i}$is the random component as described by
\begin{equation}
   \epsilon=\sqrt{1-\beta_{j}^{2}}\times Z_{i,j} 
\end{equation}
and $Z_{i,j}$ is the randomly generated normal variate with zero mean and unit variance, $\beta_{j}$is the lag-one serial correlation between month $j$ and $j-1$ given by
\begin{equation}
\beta_{j}=\frac{\sum d_{x}d_{y}}{\sqrt{\sum d_{x}^{2}\times\sum d_{y}^{2}}}
\end{equation}
where $d_x$ and $d_y$ are the derivatives of inflows in month $j$ and $j-1$ from their respective means.
A suitable starting value and the sample estimates of monthly parameters $\bar{x_j}$ and $\sigma_{j}$ are required to generate a continuous of T years of synthesized monthly flows,$x_i,_j$ where $i=1$, T and $j=1,12$ for monthly flows, Since the next period's inflow can be estimated from the previous period inflow , (9) can be used recursively to generate synthetic sequence of desired duration but it is highly dependent upon the random seed that is used to initialize set of random number in a form of pseudo random sequence. The choice of the seed might leave the Markov Chain non- stationary in some cases, which means that the reproducibility  of the results obtained from Thomas Fiering Model is less likely on day today basis. Seed given in this model is 9001.

\begin{figure}
    \centering
    \includegraphics[width=\textwidth]{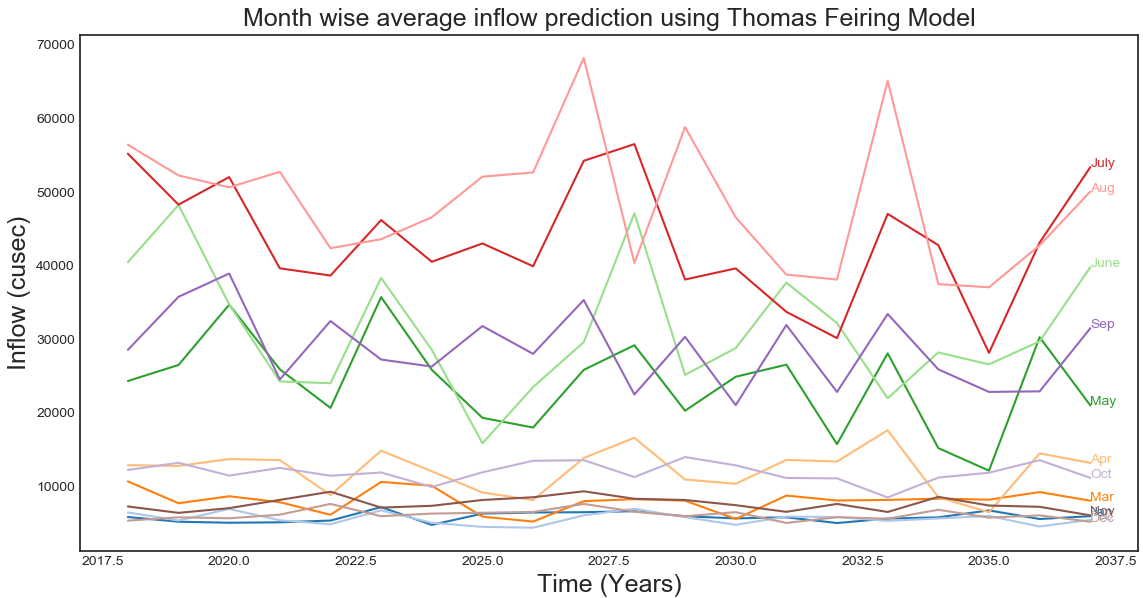}
    \caption{20 years of month wise prediction of inflow using Thomas Fiering model on Bhakra Nangal data from year 2017-2037}
    \label{fig:my_label}
\end{figure}
\section{Long Short term Memory (LSTM)}
\label{sec5}
The Long Short Term Memory \cite{hochreiter1997lstm} (Hochreiter et al, 1997) is a recurrent neural network that is trained using back propagation through time and overcomes vanishing   gradient problem. LSTM networks follow the standard computation graph model for processing continual input streams that are not a prior segmented into sub sequences. The network consists of several gates ”forget gate”, ”input gate” and ”output gate” and also in that sequence. The Equations (6-11) represent the order in which computation for each gate is done within a LSTM cell.  The key  to LSTM is that it contains a cell state $c_t$  which maintains  the state of each cell and it’s update is shown in Equation (9).
\begin{figure}
        \begin{subfigure}[b]{\textwidth}
                \includegraphics[width=\linewidth]{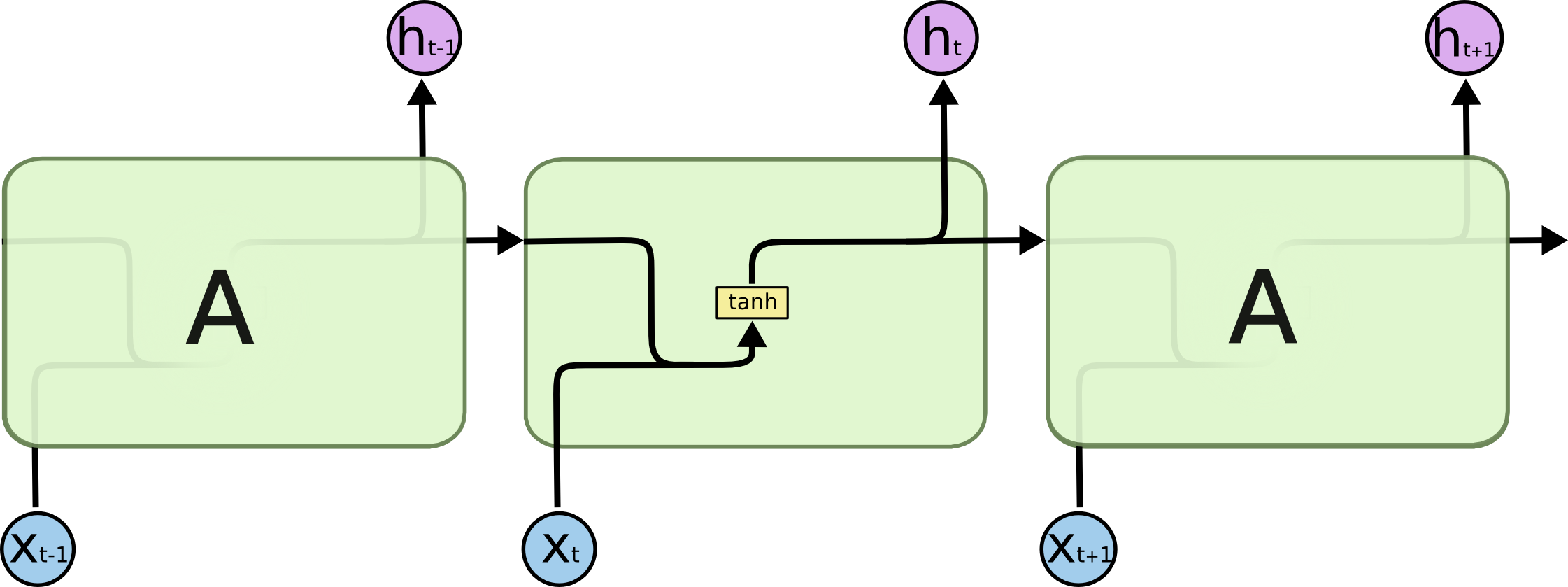}
                \caption{Recurrent Neural Networks consist of the loop within the network and is unwrapped through time in the above figure to explain the flow of input and gradients in the model. The simple model consists of an tanh activation at heart that takes current input and output.}
                \label{fig:RNN}
        \end{subfigure}
        \begin{subfigure}[b]{\textwidth}
                \includegraphics[width=\linewidth]{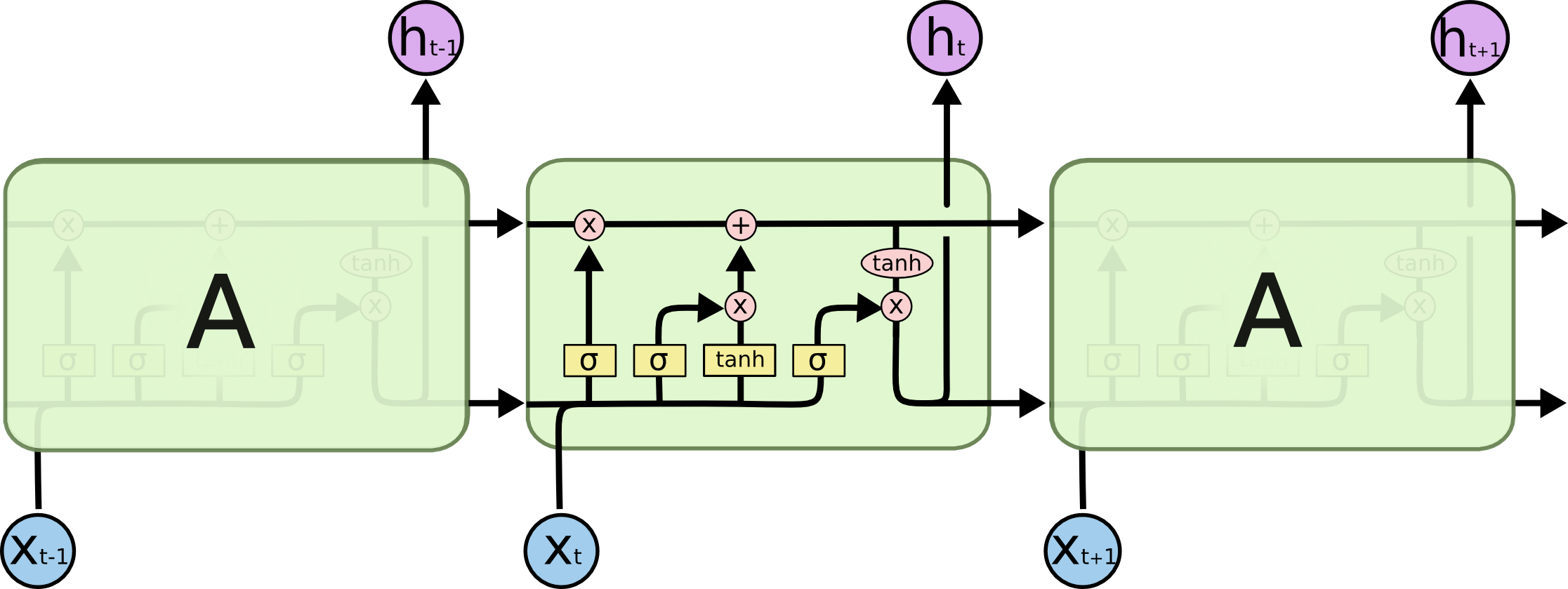}
                \caption{LSTM unwrapped through time shows how the model is designed differently from a RNN model which consists of two different loops acting as an input an output at each time step. One is the output from "output gate" and other is output from the "forget gate".}
                \label{fig:LSTM}
        \end{subfigure}
        \caption{Architectural Differences of RNN and LSTM}\label{Neural netowrk}
\end{figure}

\par
Also, it can be noticed the cell state runs through the LSTM in Fig \ref{fig:LSTM} on top and has minimal interaction with rest of the cell. The forget is described as $f_t$ which is a number in 0 to 1 and is multiplied to cell state to regulate how much of information does it need to keep. The input gate and output gate along with hidden layer computation are defined in Equations (5),(8) and (9) respectively. 
\begin{equation}
     f_{t}=\sigma(W_{f}\cdot [h_{t-1},x_t]+b_f)
\end{equation}
\begin{equation}
     i_{t}=\sigma(W_{i}\cdot [h_{t-1},x_t]+b_i)
\end{equation}
\begin{equation}
     \hat{c}_{t}=\sigma(W_{\hat{c}}\cdot [h_{t-1},x_t]+b_{\hat{c}})
\end{equation}
\begin{equation}
    c_t = f_t\odot c_t-1 + i_t\odot \hat{c}_t
\end{equation}
\begin{equation}
    o_{t}=\sigma(W_{o}\cdot [h_{t-1},x_t]+b_o)
\end{equation}
 \begin{equation}
    h_{t}=o_{t}\odot\phi(c_{t})
\end{equation}
Its design is such that it enables an LSTM cell to learn to reset itself at appropriate times, thus releasing internal resources or gradients. The novelty of "forget gate" is that it locks the gradient which helps in remembering long term dependencies of the input data. Therefore the information from the forget gate is not propagated back in time. Thus it helps to remove the vanishing gradient descent problem that the Recurrent Neural Networks(RNNs) faces and is the reason why RNNs tend to perform poorly to keep in memory and exploit temporal relationships that extend over large time steps.LSTM with forget gates, however, easily solve them in an elegant way as compared to other sequential models such as RNNs and Hidden Markov models. 
\section{Experiments}
\begin{figure}
    \centering
    \includegraphics[width=0.3\textwidth,height=120px]{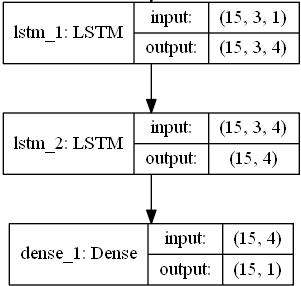}
    \caption{Architecture of Stacked LSTM network used in experiments with input batch size of 15. The respective layers have 96, 144 and 5 trainable parameters including both weights and biases. There are total of 245 trainable and 0 non trainable parameters in the network.}
    \label{fig:5}
\end{figure}

To test the efficacy of the proposed model multiple experiments are run for evaluation at training and testing time. An effort is made to keep the empirical evidence transparent and reproducible, implementation details i.e. code and data are made public\footnote{https://github.com/Anurag14/Inflow-Prediction-Bhakra}. Figure\ref{fig:5} describes visualization of the network with the help of open libraries(keras and graph viz). We use a variant of LSTM with two densely packed layers. The first entry i.e. 15 in each of tuples across all  the layers of network denotes the batch size. It is the size of input data for which network runs predictions in parallel. It helps in leveraging graphic processing unit (GPU) for faster computations and training of network. Network can be designed with different batch sizes ranging from 1 to length of data. We use batch size of 15 because it perfectly divides both our train and test datasets and helps us in faster computation by leveraging GPU. The  second number in the tuple i.e. 3 in input is look back. In other words, it is the number of past days needed to make prediction for coming day. Similar to batch size, look back can  be adjusted in the parameters fed to the LSTM while training. For all the experiments and results obtained in the given section look back of 3 days was considered.
\subsubsection{Training and RMSE}
\begin{figure*}
\centering
         \begin{subfigure}[t]{0.5\textwidth}
                \centering
                \includegraphics[width=0.8\linewidth]{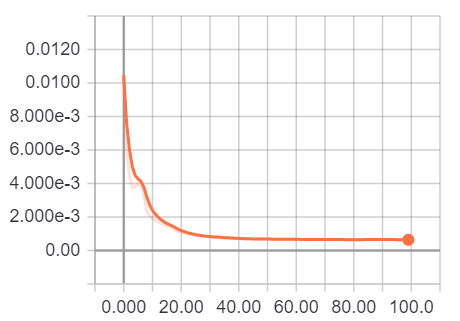}
                \caption{Train Loss}
                \label{trainloss}
        \end{subfigure}%
        \begin{subfigure}[t]{0.5\textwidth}
                \includegraphics[width=0.8\linewidth]{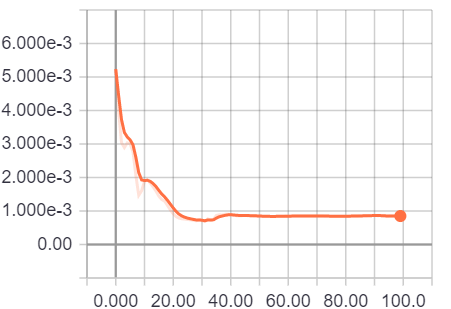}
                \caption{Validation Loss}
                \label{valloss}
        \end{subfigure}
        \caption{Training and Validation loss plots for 100 epochs}\label{train_and_val_loss}
\end{figure*}

\begin{figure}
    \centering
    \includegraphics[width=\textwidth,height=250px]{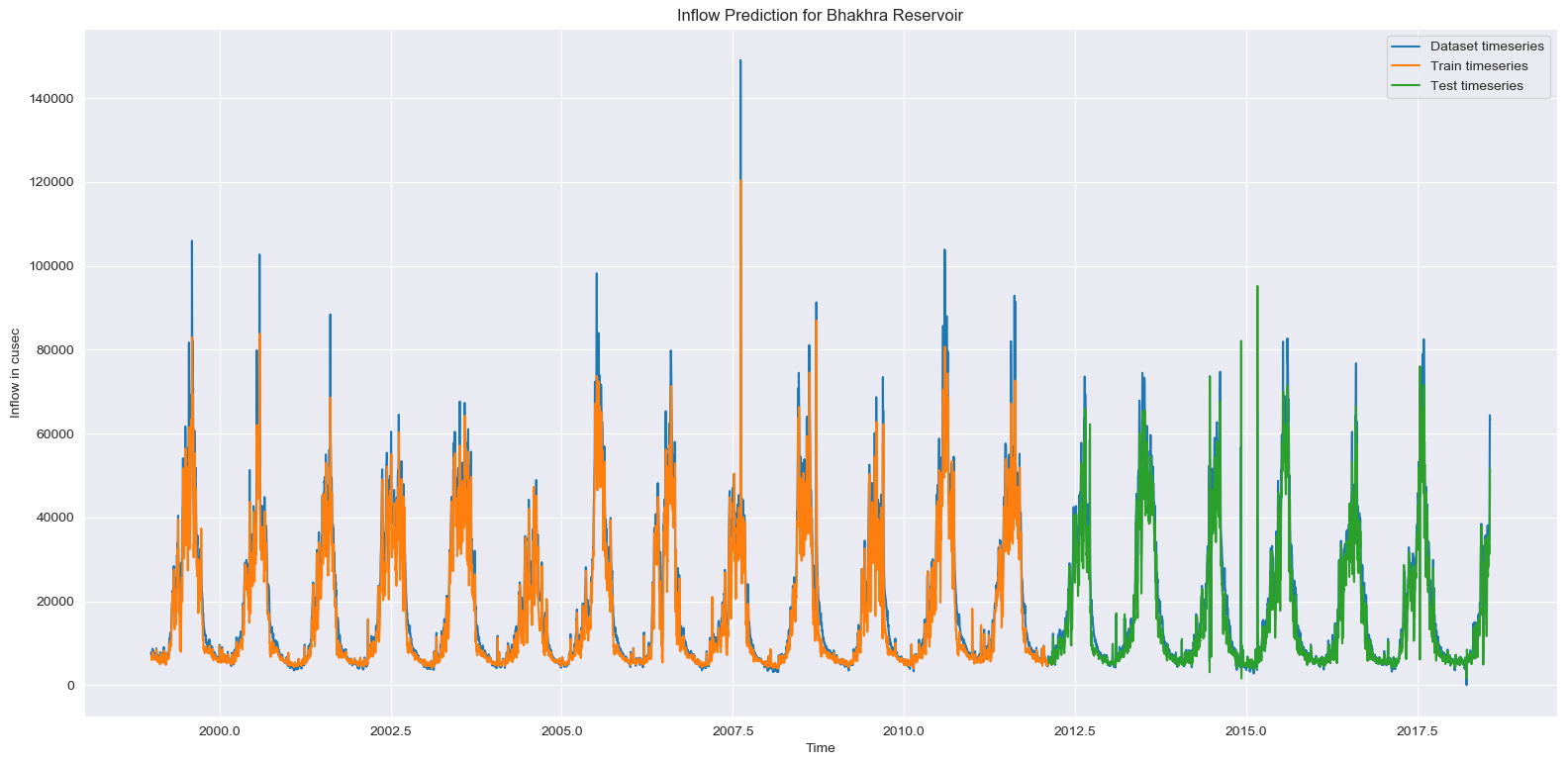}
    \caption{Time series plot of original dataset and train and test }
    \label{fig:training_plot}
\end{figure}

The Stacked LSTM is trained using tensorflow framework\cite{abadi2016tensorflow} on a NVIDIA 920 M GPU for ~100 epochs with batch size of 15 throughout. Standard Mean square Error(MSE) is used as a loss to train the model to fit the data. The model is trained on past data of 20 years of inflow data at Bhakra Dam obtained from Bhakra Beas Management Board. Precisely two thirds of which, i.e. first 13 years of data is used for training. One tenth of the data from first 13 years of training set i.e. last one year is kept aside as validation set for validation after every 5 epochs. The rest of data is never shown to the model — 7 years of latest data upon which we make inference and test our model. Model is constantly tested on test data after every 10 epoch, however, it doesn't learn from test data. Figure \ref{fig:training_plot} describes the time series view of observed data in blue vs predictions on training data in orange and test data in green. It can be observed that the predictions of the model after ~100 epochs tightly follow the trends observed in original data making the time series generated by the model. To quantify the difference between predictions made by the model and to compare with existing methods of inflow prediction we use Root Mean Square Error(RMSE) and $R^2$ as a metrics to evaluate the error correlation of predictions made by the model with ground truth. The input inflow time series that is fed to the LSTM network is first normalized between 0 to 1. The standard normalization is done using min max normalization. 
\begin{equation}
    \hat{\mathcal{X}}_{i}=\frac{\mathcal{X}_{i}-\min\limits_{\forall i}\mathcal{X}_{i}}{\max\limits_{\forall i}\mathcal{X}_{i}-\min\limits_{\forall i}\mathcal{X}_{i}}
\end{equation}
The root mean squared values are also calculated for the normalized input throughout in the experiments. The mean squared error also acts as the loss for the training of model. The loss on training and validation set vs the number of epochs is shown in Figure \ref{train_and_val_loss}. 
\begin{table}[]
\centering
\begin{tabular}{| *{9}{c|}}
\hline
\multicolumn{2}{|c|}{} & \multicolumn{7}{c|}{Epochs} \\
\cline{3-9}
\multicolumn{2}{|c|}{Dataset}&1 & 10 & 20 & 30 & 40 & 50 & 100 \\
\hline
\multirow{2}{*}{Test}& RMSE & 0.11 & 0.04 & 0.04 & 0.03 & 0.03 & 0.03 & 0.03\\
& $R^2$ & 0.9165 & 0.9256 & 0.9215 & 0.9193 & 0.9170 & 0.9140 &0.9053\\
\hline 
\multirow{2}{*}{Train}& RMSE & 0.11 & 0.04 & 0.03 & 0.03 & 0.03 & 0.03 & 0.03\\
& $R^2$ & 0.9027 & 0.9520 & 0.9507 & 0.9489 & 0.9487& 0.9492 & 0.9434\\
\hline
\end{tabular}
\caption{Epoch wise RMSE and $R^2$ of stacked LSTM Model for train and test data} 
\label{tab:table1}
\end{table}
In table \ref{tab:table1} the RMSE and ($R^2$) for train set and test set is given. It can be noticed that the RMSE performance for both test and train set improves as the model learns. Where it can also be observed that the model learns decent parameters and after one full epoch of learning is able to make predictions on trains set with RMSE of 0.11 and on test set with RMSE 0.11. The learnable parameters are kept simple in the LSTM model, explained in Figure \ref{fig:5} i.e. 245 total parameters, which shows that if more powerful network along with more data can result in even better predictions. LSTM based neural network architecture proposed in this paper is shown to have RMSE of 0.03 on test and train sets from the daily inflow data of 20 years. Table \ref{comparison_table} is comparison of our approach with other acceptable baselines on recently sourced inflow data of 2018-2019.

\begin{table}[]
\centering
 \begin{tabular}{|l|c|c|}
 \hline
  \multirow{2}{*}{Evaluation for $1^{st}$ May, 2018 to $30^{th}$ April, 2019} & \multicolumn{2}{c|}{Metrics} \\
  \cline{2-3}
   & RMSE & $R^2$ \\
  \hline
  Thomas Fiering Monthly &0.1207 & 0.8933\\
  \hline
  Thomas Fiering Daily &0.1420 & 0.6766\\
  \hline
  LSTM Daily & \textbf{0.0503} & \textbf{0.9389}\\
  \hline
  10 Daily & 0.2940 & 0.6571 \\
 \hline
 \end{tabular}
 \caption{RMSE and $R^2$ Values for Monthly  and Daily Inflow forecast results }
 \label{comparison_table}
 \end{table}
Thomas Fiering model is a the standard model which is used for inflow determination and is used as the baseline for comparison in our further experiments. Since Thomas Fiering model just uses lag-1 auto-correlation it tends to estimate the presence of any trend based on observation of just one previous entry. In table \ref{comparison_table} it can be observed that the monthly average estimates made using Thomas-Fiering model tend to give a RMSE of 0.1207  and ($R^2$) 0.8933. The Thomas-Fiering model is also extended to give daily predictions for comparisons. When compared for one year duration it gives RMSE value 0.1420 and ($R^2$) value 0.6766. LSTM give RMSE 0.0503 and ($R^2$) 0.9389. It can therefore be concluded that the Thomas Fiering model is not very suitable for very accurate daily predictions of inflow in a reservoir. The daily inflow bears more significance for real time operations of the reservoir and its strategies as compared to monthly average inflow. 
\section{Anomaly Detection}
\label{7}
Inflow prediction for reservoir operations helps in making better and more sound real time micro management strategies. Daily inflow prediction with certain accuracy can help in creating better awareness regarding what inflow to expect on normal days. However, in certain scenarios there might be instances where the predictions are very different from the observed values. The model is trained on cases borrowing data from large corpus of past it learns parameters and generalizes. It has to be robust to anomalies or outliers to be good at making general predictions. There are several techniques of regularization used in machine learning community to prevent the model from over-fitting to such anomalies as it leads to poorer predictions in real world. Having said that, there are sometimes cases where detection of anomalies is also of great significance. One such use case is flood and drought prediction. 

\begin{algorithm}
\caption{Naive Anomaly Detection based on LSTM}\label{anamoly}
\begin{algorithmic}[1]
\Procedure{PredictFloodOrDrought}{$LSTM$,$lookback$,$groundtruth$,$k$,$\rho$,$\tau=0.03$}
\State $Observations \gets NormalizeObservations(groundtruth)[-k:]$
\Comment{last $k$ days}
\State $input \gets Observations[-(k+lookback):-k]$ \Comment{Take lookback entries before $k$}
\State $Predictions \gets null$
\For{$k$ iterations}
\State $Prediction \gets LSTM(input)$
\State $Input \gets Input[1:]$ \Comment{Remove of last entry}
\State $Input.append(Prediction)$ \Comment{Insert Prediction}
\State $Predictions.append(Prediction)$
\EndFor
\State $ObservedRMSE \gets RMSE(Predictions,Observations)$
\If {$ObserverdRMSE > \tau\rho$} 
\State \textbf{Anomaly is observed}
\State $ObservedInflow,PredictedInflow \gets sum(observations),sum(predictions)$ 
\If {$PredictedInflow <ObservedInflow$} 
\State \Return \textbf{Flood}
\Else 
\State \Return \textbf{Drought}
\EndIf
\EndIf
\EndProcedure
\end{algorithmic}
\end{algorithm}
The  below Algorithm tries to describe the proposed naive algorithm baseline using LSTM models. This naive algorithm baseline can be extended to any deep learning model in future as in nutshell the algorithm compares the predictions on previous $k$ days with the observed inflow values. Before that, the ground truth that is the absolute observed values are first normalized using min max normalization. The comparison is based upon the RMSE values obtained upon the normalized data. Predicted values give a ball park figure of what the normal inflow should have been. $\tau$ is the empirical RMSE value that is derived using the experiments of training LSTM model on past ~20 years of data. Hence, it is safely assumed that the RMSE of the observed value with current prediction must remain in certain tolerance with the empirical RMSE value i.e. $\tau$. Now, that tolerance is defined as $\tau\rho$ where $\rho$ is a multiplier and $\rho>1$. For safely concluding that the observed inflow trend is anomalous $\rho=2$ can be taken but in general $\rho$ must be a linear multiplier i.e. tolerance must be $\mathcal{O}(\tau)$.  
\section{Major contributions of this paper are:}
\label{9}
A LSTM based deep neural network architecture has been used for reservoir daily inflow forecasts. 
\begin{itemize}
    \item Multiple experiments are conducted to prove efficacy of LSTM in calculating daily inflow levels both by qualitative measure such as Figure \ref{fig:training_plot} and quantitatively as comparison of Root Mean Square Errors and coefficient of determination $R^2$ between ground truth and daily predictions from LSTM, Thomas-Fiering and 10 Daily procedure.
    \item RSME and R2 for widely accepted Thomas-Fiering model for monthly average inflow prediction with monthly average observed inflow are also provided as reference.
    \item A naive algorithm baseline for anomaly detection i.e. flood and drought prediction based upon LSTM is also proposed.
\end{itemize}
It is suggested release policy can be more robust if they operate dam considering LSTM daily inflow predictions into account.
\begin{equation}
    Daily  \hspace{1mm}release  \hspace{1mm}from  \hspace{1mm}storage =\frac{Available\hspace{1mm} storage }
       {Total \hspace{1mm}water \hspace{1mm}indent\hspace{1mm} during\hspace{1mm}remaining \hspace{1mm}year}
\end{equation}

\begin{equation}
    Total\hspace{1mm} Daily \hspace{1mm}release = Daily \hspace{1mm}release \hspace{1mm}from\hspace{1mm} storage + Inflow \hspace{1mm}predicted\hspace{1mm} using\hspace{1mm} LSTM
\end{equation}
Quantity of water available for daily discharge depends upon:
\begin{itemize}
    \item Daily release possible from available storage depends on the storage in the reservoir at the end of filling period.
    \item Daily inflow forecasted using LSTM with 3\% RMSE.
\end{itemize}
Use of above criteria  for release of water will reduce the error for inflow determination currently followed from 29.4\% to 3\% using LSTM code, providing better decision policy for reservoir operation 
\section{Conclusion}
\label{10}
With population growth increase in requirement of power and reliable water supply increases. With increase in demand, more clever reservoir operation is needed to meet power demands and provide reliable water supply. With the use of machine learning techniques wide range of problems in optimization and operations research are being solved. Deep learning has become ubiquitous in recent state of art solutions that are being employed to wide array of tasks. An attempt has been made in this study for determination of inflow at Bhakra Dam. 
Daily inflow predictions using LSTM will enhance operational performance of reservoir. Model is able to predict daily inflow with Root Mean Square error of 3\% and $R^2$ value 0.9389. Thomas-Fiering model is also extended from monthly prediction to daily prediction. The Root Mean Square error is 14.20\% and $R^2$ value is 0.6766 compared to observed values. Using Thomas- Fiering Model monthly prediction were also made and results when compared with observed monthly resulted in Root Mean Square error as 12.07\% and $R^2$ value as 0.8923. RMSE and $R^2$ values clearly indicate that application of LSTM  in reservoir operation and daily prediction of inflow  performs best compared to prevailing technique followed at Bhakra dam and Thomas- Fiering Model predictions. Using LSTM release from the reservoir can be monitored efficiently. Use of daily inflow prediction in estimating the release will reduce the complexities of reservoir operation. Use of monthly inflow overlooks daily variation of inflow in making decision regarding releases which can be avoided. Naive Anomaly detection based on LSTM will be able to warn water management regarding flood and drought. If the value of RMSE shows large variation between observed inflow and LSTM predicted inflow, it is clear indication of flood or drought. Operational release policy can be updated accordingly in order to have efficient operation. Daily inflow determination  modeling technique is expected  to improve the reservoir operational strategies under changing climatic conditions. Streamflow generation model are used to synthesize daily inflow sequences considering previous inflow. Comparison between synthetic data from these models and observed data in terms of RMSE can help evaluate climate change.

\appendix


\bibliographystyle{elsarticle-num}

\bibliography{sample}

\end{document}